\documentclass[conference]{IEEEtran}
\IEEEoverridecommandlockouts
\usepackage{cite}
\usepackage{amsmath,amssymb,amsfonts}
\usepackage{algorithmic}
\usepackage{graphicx}
\usepackage{multirow}
\usepackage{textcomp}
\usepackage{xcolor}
\usepackage{subcaption}
\usepackage{adjustbox}
\renewcommand{\thefootnote}{\fnsymbol{footnote}}

\def\BibTeX{{\rm B\kern-.05em{\sc i\kern-.025em b}\kern-.08em
    T\kern-.1667em\lower.7ex\hbox{E}\kern-.125emX}}
\begin{document}

\title{Multimodal Fusion with BERT and Attention Mechanism for Fake News Detection}

\author{\IEEEauthorblockN{Nguyen Manh Duc Tuan$^*$ \thanks{$^*$This work was done when the author was an internship student at Aimesoft JSC}}
\IEEEauthorblockA{\textit{Toyo University}\\
Tokyo, Japan \\
ductuan024@gmail.com}
\and
\IEEEauthorblockN{Pham Quang Nhat Minh}
\IEEEauthorblockA{\textit{Aimesoft JSC}\\
Hanoi, Vietnam \\
minhpham@aimesoft.com}
}

\maketitle
\renewcommand{\thefootnote}{\arabic{footnote}}

\begin{abstract}
Fake news detection is an important task for increasing the credibility of information on the media since fake news is constantly spreading on social media every day and it is a very serious concern in our society. Fake news is usually created by manipulating images, texts, and videos. In this paper, we present a novel method for detecting fake news by fusing multimodal features derived from textual and visual data. Specifically, we used a pre-trained BERT model to learn text features and a VGG-19 model pre-trained on the ImageNet dataset to extract image features. We proposed a scale-dot product attention mechanism to capture the relationship between text features and visual features. Experimental results showed that our approach performs better than the current state-of-the-art method on a public Twitter dataset by 3.1\% accuracy.
\end{abstract}


\section{Introduction}

Recently, fake news detection has received much attention in both NLP and the data mining research community. In earlier research, any kind of content such as satires, hoaxes, or click baits was considered fake news. However, Shu et al.~\cite{shu2017} has defined fake news as ``news articles that are intentionally and verifiably false, and could mislead readers''. An example of how fake news can affect us is the 2016 U.S. presidential elections. In the final three months before the election, fake news favoring either of the two nominees was shared by more than 37 million times on Facebook~\cite{Allcott16}. That makes detecting fake news be a necessary task.

\begin{figure}[!ht]
    \centering
    \includegraphics[width=0.6\linewidth]{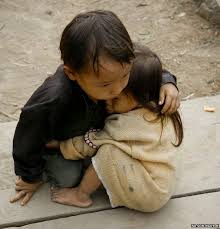}
    \caption{\textbf{Examples of misleading image use: reposting of real photo claiming to show two Vietnamese siblings at Nepal 2015 earthquake}}
    \label{fake}
\end{figure}

Our work is inspired by the idea that different modalities can show different aspects of news and different modalities can complement each other in detecting the authenticity of news~\cite{SpotFake}. Fake news is usually created by both images and texts. In fake news, images may not represent the event that the text content refers to~\cite{mediaeval16}. Figure~\ref{fake} is an image of two kids and they may be siblings, but the image is neither about \textit{Vietnamese people} nor \textit{the earthquake in Nepal in 2015}.

In this paper, we present a novel multimodal model for identifying fake news. We use neural networks to obtain feature representations from different modalities. Multimodal features are fused using the attention mechanism and put into a sigmoid layer for classification. Specifically, we use the BERTweet model~\cite{bertweet} to obtain feature representations from texts and a pre-trained VGG-19 network to obtain feature representations from images. We propose a scaled dot-product attention mechanism on both texts and images, and also a self-attention mechanism on images because we see that in non-photoshopped images, all parts of images are related. Textual representations and visual representations along with three attention outputs are combined to improve the accuracy of fake news detection. Our proposed model obtained 80.8\% of accuracy and 80\% of F1-score.

The rest of the paper is organized as follows. Section~\ref{related} discusses the existing work on fake news detection, especially studies done using multimodal architectures. In Section~\ref{sec:method}, we present our proposed method for fake news detection and in Section~\ref{sec:dataset}, we describe the dataset that we used in experiments. Section~\ref{sec:experiment} gives our experimental results and results analysis. Finally, we conclude the paper in Section~\ref{sec:conclusion}.

\section{Related work}
\label{related}

The majority of previous work on fake news detection used text and user metadata features. Specifically, the textual features can be obtained by applying convolutional neural networks (CNN)~\cite{kaliyar2020fndnet,tian2020stance}. Conneau et al.~\cite{conneau2017deep} has shown that a deep stack of local operations can help a model to learn the high-level hierarchical representation of a sentence and that increasing the depth leads to the performance improvement. Furthermore, deeper CNNs with residual connections can help to avoid over-fitting and solve the vanishing gradient problem ~\cite{kaliyar2020fndnet}. Textual features can also be manually designed from word clues, patterns, or other linguistic features of texts such as their writing styles~\cite{ghosh2018towards,Wang2018EANNEA,yang2018ticnn}. We can also analyze unreliable news based on the sentiment analysis ~\cite{8508256}. 
 
Fake news can be detected by analyzing social network information including user-based features and network-based features. User-based features were extracted from user profiles in~\cite{shu2019role,8424744,duan2020rmit}. For example, the number of followers, number of friends, and registration ages are useful features to determine the credibility of a user post~\cite{Castillo2011}. Network-based features can be extracted from the propagation of posts or tweets on graphs~\cite{zhou2019networkbased,ma-etal-2018-rumor}.

Recent studies have shown that visual features and correlations between modalities are useful factors in detecting fake news~\cite{Wu2015FalseRD,zhou2020safe,Khattar2019,yang2018ticnn,jin-att-rnn,SpotFake}. Many of them have used word embeddings which is low-level embeddings for getting textual representations. Different from those, we applied the BERTweet model for extracting sentence embeddings.

Jin et al.~\cite{jin-att-rnn} created an end-to-end network called att-RNN. It used attention mechanisms to combine textual, visual, and social context features. In that network, texts and social contexts were concatenated together and passed into an LSTM network. Image features were extracted from a pre-trained VGG-19 model and the outputs of the LSTM are used with a attention mechanism for fusing the visual features. Finally, the average of outputs of the LSTM was joint with the visual features to make prediction.

Wang et al.~\cite{Wang2018EANNEA} built an end-to-end model called Event Adversarial Neural Networks for Multi-Modal Fake News Detection (EANN). That model has two components: event discriminator and fake news classification. For an input text, they used word embedding vectors as input and used CNN layers on it to generate the textual representation. The representation for an image was extracted from an VGG-19 model pre-trained on the ImageNet dataset. Finally, both of these representations were concatenated and passed into two fully connected layers on top of the above two components.

Khattar et al.~\cite{Khattar2019} built a similar architecture called Multimodal Variational Autoencoder for Fake News Detection (MVAE). Similar to EANN, they used word embedding vectors, but they used bi-directional LSTMs to extract text representations. Image representations were also extracted from a VGG-19 pretrained model. The latent vectors were created by concatenating these two vectors and were passed into a decoder for reconstructing original samples. The latent vectors are then passed into two fully connected layers for fake news detection.

Singhal et al.~\cite{SpotFake} created a survey to show the essence of texts and images in fake news detection and they built an architecture called Spotfake. It uses BERT to obtain text feature representations and a pre-trained VGG-19 network to obtain feature representations from images. Then two single models were combined to make prediction. While that method obtained good results, it did not make use of the correlation between modalities, which is useful for fake news detection. In the survey, Singhal et al.~\cite{SpotFake} showed that 81.4\% of the people, who took the survey, were able to distinguish fake news from real news when both images and texts are given, while the number is 38.4\% if only texts are given and 32.6\%  if only images are given. That indicated that multiple modalities provide more information and useful for fake news detection.

\section{Methodology}
\label{sec:method}

\begin{figure*}[h!]
    \centering
    \includegraphics[width=0.67\linewidth]{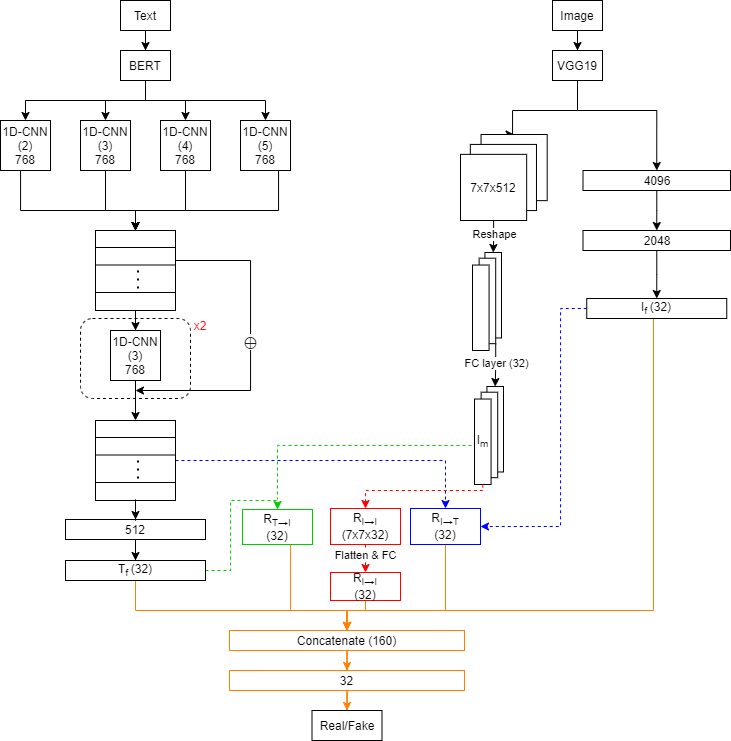}
    \caption{Model Architecture}
    \label{fig:model}
\end{figure*}

In this section, we describe our multimodal approach to fake news detection. Figure~\ref{fig:model} shows the architecture of our proposed model. The model contains four parts. The first part is the textual feature extractor in which we use BERT~\cite{devlin2019bert} and CNN layers to extract the contextual text features. The second part of the model is the visual feature extractor that extracts the visual features from an image in a post. The third part is the common feature extractor in which we proposed an attention mechanism to extract the features from both texts and images. Finally, the last part is a multiple feature combination component which merges the representations derived from different components to obtain the feature representation for the entire post. 

\subsection{Textual Feature Extractor}

Before extracting textual features, we performed the following pre-processing steps on texts.

\begin{itemize}
    \item We converted words and tokens that have been lengthened into short forms. For example, ``Coooool'' into ``Cool''. 
    \item There are some emojis written in text format such as ``:)'', ``:('', etc. We changed those emojis into sentiment words ``happy'' or ``sad''.
    \item We did tokenization, word normalization, word segmentation with ekphrasis~\cite{baziotis-pelekis-doulkeridis:2017:SemEval2}, a text analysis tool for social media.
\end{itemize}

We use BERT, the state-of-the-art model in many NLP tasks, to extract the feature representation of a tweet in a way that best captures underlying semantic and contextual information. BERT model has been shown to be effective in many NLP tasks including text classification. In this task, We use BERTweet~\cite{bertweet}, a BERT model pre-trained on Tweet data. Cheema et al.~\cite{cheema2021tibs} and Devlin et al.~\cite{devlin2019bert} have suggested that different hidden layers of BERT can capture different kinds of information of the text, and the last four hidden layers of BERT are good for extracting information in a feature-based approach. Thus, we concatenate the last 4 hidden layers as contextual embeddings of tokens. After that, we use 1D-CNN layers~\cite{kim2014convolutional} with filter sizes 2, 3, 4, and 5 to extract more information from different sets of word vectors for prediction. After passing the embedding vectors through 1D-CNN layers, we stack those outputs vertically and pass into two additional 1D-CNN layers with residual connections and obtain the output \(T_m\) in which the number of filters in each CNN layer is \(d_{T}=768\). Finally, we flatten \(T_m\), pass it through two fully connected layers and obtain \(d_{T}=32\) as the final vector size of the textual representation \(T_f\).

\subsection{Visual Feature Extractor}

We use an VGG-19 model pre-trained on ImageNet dataset~\cite{simonyan2015deep} for visual feature extraction. We extract the output of the second last layer of VGG-19 and pass it through 2 fully connected layers to reduce the dimension to a vector size \(d_{I}=32\) as the final visual representation \(I_f\). We also extract the third last layer output \(I_m\) that we will use later in the common feature extracting part. \(I_m\) is reshaped from 3D tensor to 2D tensor of shape \(regions \times d_{I_m}\) where $regions$ is the number of pixels in the output of the second last layer of VGG-19. There are \(7 \times 7 = 49\) regions for an image of size \(224 \times 224\). Finally, we put it into a fully connected layer and obtain the visual feature representation of size \(d_{I_m}=32\).

\subsection{Common Feature Extractor}

\begin{figure}[!ht]
    \centering
    \includegraphics[width=0.6\linewidth]{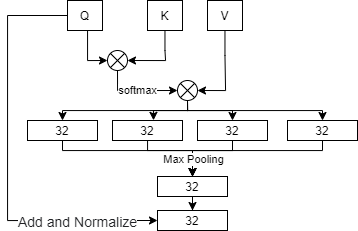}
    \caption{Scaled-dot product attention mechanism}
    \label{attention}
\end{figure}

In this part, we proposed to apply the scaled dot-product attention mechanism on visual features and textual features (\(I_f\), \(I_m\), \(T_f\), \(T_m\)) to capture how well text and images are related to each other in a post. We applied the attention mechanism for text and visual features in two directions and we also used the self-attention mechanism on the images since we believe that contents in an unmanipulated image should relate to each other. Figure~\ref{attention} shows the general design of our proposed attention mechanism.

When we use information from a text to make the comparison to an image or we use the text vector representation \(T_f\) as Query and image regions' features \(I_m\) as Key and Value, the three terms Query, Key and Value are defined as follows.

\begin{align*}
Q = T_f \times W_Q, K = I_m \times W_K, V = I_m \times W_V\\ 
W_Q \in R^{d_T\times d}, W_K \in R^{d_{I_m}\times d}, W_V \in R^{d_{I_m}\times d}
\end{align*}

\noindent where \(d=32\), \(W_Q, W_K, W_V\) are weight matrices and \(\times\) is matrix-multiplication operation.

The output matrix of the scale dot-product attention applied on Q, K, and V is calculated as follows.

\begin{align*}
Att_{T \rightarrow I} = softmax( \frac{Q \times K^T}{\sqrt{d}}) \times V 
\end{align*}
where \(Att_{T \rightarrow I}\) is the attention's output matrix when we use text vector representation \(T_f\) as Query and image regions' features \(I_m\) as Key and Value.

Similarly, when we use information from the image to make comparison to the text, we obtain \(Att_{I \rightarrow T}\), the attention's output matrix in which the visual vector representation \(I_f\) is used as Query and \(T_m\) is used as Key and Value. 

The output matrix \(Att_{I \rightarrow I}\) of self-attention mechanism applied on an image is calculated as follows.

\begin{align*}
Q = I_m \times W_Q, K = I_m \times W_K, V = I_m \times W_V \\ 
W_Q \in R^{d_{I_m}\times d}, W_K \in R^{d_{I_m}\times d}, W_V \in R^{d_{I_m}\times d}
\end{align*}

After obtaining three attention's output matrices \(Att_{T \rightarrow I}\), \(Att_{I \rightarrow T}\) and \(Att_{I \rightarrow I}\), we pass each of them into four different fully connected layers with the size of 32, which is the same as \(d_T\), \(d_I\) and \(d_{I_m}\). Then we take a maximum of 4 vectors, pass it into another fully connected layer and add a residual connection into it. Finally, we obtain three vectors \(R_{T \rightarrow I}\), \(R_{I \rightarrow T}\) and \(R_{I \rightarrow I}\). We also use layer normalization~\cite{ba2016layer} to the output of each attention block.

\subsection{Multiple Feature Combination}

In this step, \(R_{I \rightarrow I}\) is flattened and pass into a fully connected layer of size 32 and we obtain \(R_{I \rightarrow I}'\). Finally, we concatenate 5 outputs: \(T_f\), \(I_f\), \(R_{T \rightarrow I}\), \(R_{I \rightarrow T}\) and \(R_{I \rightarrow I}'\) and pass the concatenated tensor through a fully connected layer with 32 neural units. In summary, we have 2 vector representations for textual features, which are \(T_f\),  \(R_{I \rightarrow T}\) and 3 vector representations for visual features, which are \(I_f\), \(R_{T \rightarrow I}\), \(R_{I \rightarrow I}'\).

\section{Dataset}
\label{sec:dataset}


We evaluated our proposed model on MediaEval 2016 data. The dataset was released for the Verifying Multimedia Use challenge at MediaEval 2016~\cite{boididou2018detection,mediaeval16} and is publicly available\footnote{https://github.com/MKLab-ITI/image-verification-corpus/tree/master/mediaeval2016}. The purpose of the challenge is to distinguish whether a piece of information is true/fake news. The dataset contains tweets and their associated images. It consists of 17,000 unique tweets related to various events. The training set consists of 9,000 fake-news tweets and 6,000 real-news tweets, and the test set contains 2,000 news tweets. Some tweets have attached videos, but we only kept the data with texts and attached images and removed the samples with attached videos.

\section{Experiments and Results}
\label{sec:experiment}
\subsection{Experimental setup}

In experiments, we used one image in a post as the input for visual feature extraction. We randomly chose one image from the post if the post contains multiple images.

We used BERTweet model~\cite{bertweet} for text feature extraction. We chose the maximum sequence length of 32 for the BERT model and used padding for texts whose lengths are shorter or longer than 32. In our proposed model, we kept weights of pre-trained BERT and VGG-19 fixed and used them as feature extractors, because in preliminary experiments, we found that fine-tuning BERT and VGG-19 did not improve the performance of our model.

Hyper-parameters used in experiments are as follows. The number of filters of each 1D-CNN layer is 768, and we used the pooling size of 3 in max-pooling layers. The hidden size of the fully connected layer before \(T_f\) is 768.

We resized all the images to 224x224x3. The second last output of VGG-19 has the size of 4096 and the third last output has the shape of 7x7x512. The hidden size of the fully connected layer before \(I_f\) is 2048 and the hidden size of the layer before \(I_m\) is 32.

After each fully connected layer in our proposed model, we applied the Dropout technique~\cite{JMLR:v15:srivastava14a} and set dropout rate to 0.3. We trained the model with 10 epochs, batch size of 256, and the Adam optimizer with the learning rate of 1e-4. 

\subsection{Results}

In experiments, we compared our proposed model with some baseline uni-modal models and multimodal models as follows. Baseline multimodal models have been described in Section~\ref{related}.

\begin{itemize}
    \item \textbf{TextLSTM} is an LSTM network that includes a bidirectional LSTM layer, a feed-forward layer, and a softmax layer. We used Google pre-trained word embeddings with the dimension of 32. The model used only textual features.
    \item \textbf{Textual (BERTweet)} used only textual features derived from the pre-trained BERTweet model. The model corresponds to the left component in Figure~\ref{fig:model}.
    \item \textbf{Visual} used only visual features derived by VGG-19. The model corresponds to the right component in Figure~\ref{fig:model}.
    \item \textbf{att-RNN} is the multimodal model with attention mechanism presented in~\cite{jin-att-rnn}. We did not use social features in \textbf{att-RNN} for a fair comparison.
    \item \textbf{EANN} is the Event Adversarial Neural Networks presented in~\cite{Wang2018EANNEA}.
    \item \textbf{MVAE}~\cite{Khattar2019} is a multimodal model based on Variational Autoencoder.
    \item \textbf{Spotfake}~\cite{SpotFake} combined multimodal features extracted by BERT and VGG-19.
    
\end{itemize}

\begin{table}[ht]
\Huge
    \resizebox{\columnwidth}{!}{\begin{tabular}{|c|c|c|c|c|c|c|c|}
         \hline
        \multirow{2}{*}{Model} &\multirow{2}{*}{Accuracy} & \multicolumn{3}{c|}{Fake News}    &\multicolumn{3}{c|}{Real News}  \\  
        \cline{3-8} 
         &  &{Precision} & {Recall} & {F1-score} & {Precision} & {Recall} & {F1-score}\\
         \hline 
        TextLSTM & 0.526 & 0.586 & 0.553 & 0.569 & 0.469 & 0.526 & 0.496\\
        \hline 
        Textual (BERTweet) & 0.666 & 0.667&0.840&0.743 & 0.664 & 0.430 & 0.522\\
        \hline
        Visual & 0.596 & 0.695 & 0.518 & 0.593 & 0.524 & 0.7 & 0.599\\
        \hline

        att-RNN & 0.664 & 0.749 & 0.615 & 0.676 & 0.589 & 0.728 & 0.651\\
        \hline
        EANN & 0.648 & 0.810 & 0.498 & 0.617 & 0.584 & 0.759 & 0.660\\
        \hline
        MVAE & 0.745 & 0.801 & 0.719 & 0.758 & 0.689 & \textbf{0.777} & 0.730\\
        \hline
        Spotfake & 0.777 & 0.751 & \textbf{ 0.900} & 0.82 & \textbf{0.832} & 0.606 & 0.701\\
        \hline
        Proposed model &\textbf{ 0.812} & \textbf{0.813} & 0.874 &\textbf{0.843}& 0.810 & 0.728 & \textbf{0.767}\\
         \hline
    \end{tabular}}

    \caption{Performance of proposed model vs other methods}
    \label{tab:result}
\end{table}

\begin{table}[ht]
\Huge
    \resizebox{\columnwidth}{!}{\begin{tabular}{|c|c|c|c|c|c|c|c|}
         \hline
        \multirow{2}{*}{Model} &\multirow{2}{*}{Accuracy} & \multicolumn{3}{c|}{Fake News}    &\multicolumn{3}{c|}{Real News}  \\  
        \cline{3-8} 
         &  &{Precision} & {Recall} & {F1-score} & {Precision} & {Recall} & {F1-score}\\
        \hline
        BERT base & 0.669 & 0.769 & 0.607 & 0.678 & 0.585 & \textbf{0.753} & 0.659\\
        \hline
        BERT large & 0.788 & 0.806 & 0.832 & 0.819 & 0.762 & 0.728 & 0.744\\
        \hline
        BERTweet & \textbf{0.812} & \textbf{0.813} & \textbf{0.874} &\textbf{0.843}& \textbf{0.810} & 0.728 & \textbf{0.767}\\
         \hline
    \end{tabular}}

    \caption{Performance in different version of BERT}
    \label{tab:bert}
\end{table}



\begin{figure*}[!ht]
\begin{subfigure}{.5\textwidth}
    \centering
    \includegraphics[width=0.5\linewidth]{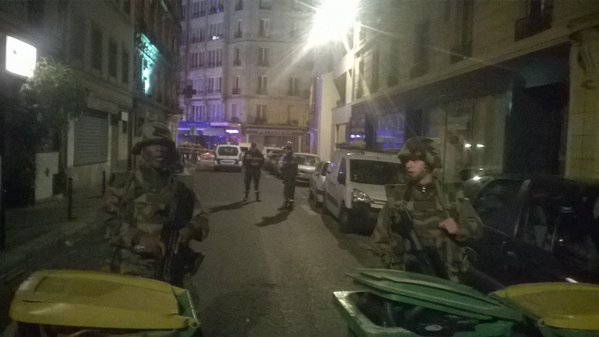}
    \caption{Real News}
    \label{fig:real}
\end{subfigure}%
\begin{subfigure}{.5\textwidth}
    \centering
    \includegraphics[width=0.4\linewidth]{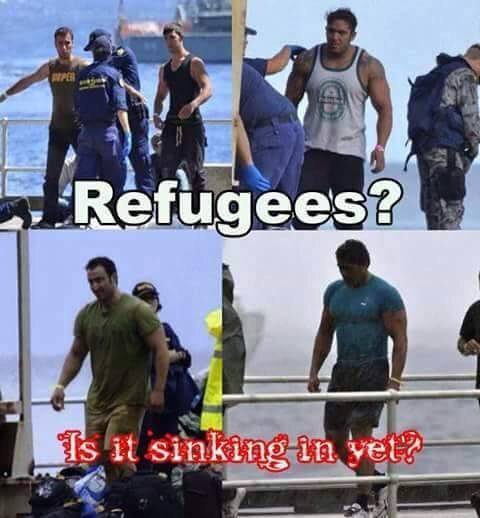}
    \caption{Fake News}
    \label{fig:fake}
\end{subfigure}
\caption{Image in the news contents.}
\label{fig:news}
\end{figure*}

\begin{figure*}[!ht]
\begin{subfigure}{.3\textwidth}
    \centering
    \includegraphics[width=0.6\linewidth]{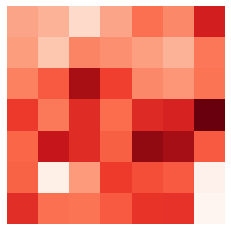}
    \caption{Text as Q and Image as K}
    \label{fig:true_text_on_image}
\end{subfigure}%
\begin{subfigure}{.3\textwidth}
    \centering
    \includegraphics[width=0.6\linewidth]{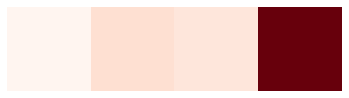}
    \caption{Image as Q and Text as K}
    \label{fig:true_image_on_text}
\end{subfigure}
\begin{subfigure}{.3\textwidth}
    \centering
    \includegraphics[width=0.6\linewidth]{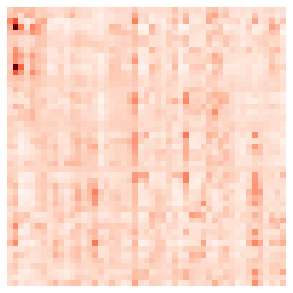}
    \caption{Self-attention on Image}
    \label{fig:true_self_attention}
\end{subfigure}
\caption{Attention outputs on real news.}
\label{fig:true_news}
\end{figure*}

\begin{figure*}[!ht]
\begin{subfigure}{.3\textwidth}
    \centering
    \includegraphics[width=0.6\linewidth]{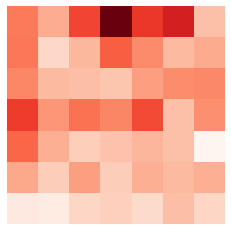}
    \caption{Text as Q and Image as K}
    \label{fig:fake_text_on_image}
\end{subfigure}%
\begin{subfigure}{.3\textwidth}
    \centering
    \includegraphics[width=0.6\linewidth]{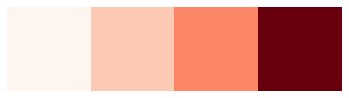}
    \caption{Image as Q and Text as K}
    \label{fig:fake_image_on_text}
\end{subfigure}
\begin{subfigure}{.3\textwidth}
    \centering
    \includegraphics[width=0.6\linewidth]{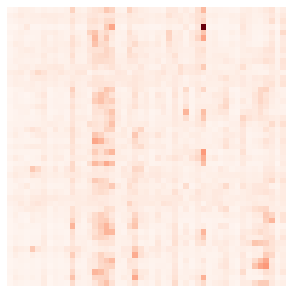}
    \caption{Self-attention on Image}
    \label{fig:fake_self_attention}
\end{subfigure}
\caption{Attention outputs on fake news.}
\label{fig:fake_news}
\end{figure*}

Table~\ref{tab:result} shows the results of the baselines and our proposed method on the MedialEval 2016 dataset. Our proposed model outperformed baseline uni-modal models and multi-modal models. In comparison with Spotfake - a model which combines multimodal features derived from pre-trained BERT model and pre-trained VGG-19 model, our model improved 3.1\% in accuracy and 4.4\% in Macro-F1.



We also compared two versions of BERT in the proposed model: BERTweet and BERT base in table~\ref{tab:bert}. The results indicated that on MediaEval 2016 data, BERTweet outperforms BERT models trained on general domain corpus. A plausible explanation is that BERTweet was trained on 850 million English Tweets which have the same domain as MediaEval 2016 data. 


In order to show the effectiveness of our proposed attention mechanism, we plot the attention weight heat maps for a piece of real news and a piece of fake news in Figure~\ref{fig:real} and Figure~\ref{fig:fake}, respectively. The real news (Figure~\ref{fig:real}) comes with text content:\textit{ RT @saserief: “@Conflicts:  French military deployed on the streets of \#Paris  large scale terror attack taking place - @lepoint https://t.…} and the fake news (Figure~\ref{fig:fake}) has the content: \textit{RT @londonorganiser: Ever get the feeling you're being had Europe? http://t.co/shIVo0S7RZ}. There is clear discrimination between the two images through the heat maps. In the fake news' image, regions are less related to each other since it is an manipulated image from 4 different images with added words. Figure~\ref{fig:true_image_on_text} shows that the picture can clearly express the content of the tweet (and vice versa), which is about \textit{military deployed on the streets}, meanwhile, there is no connection between text and image in the fake news, so it is very ambiguous when using the picture to express the content of the tweet.

\section{Conclusion}
\label{sec:conclusion}

We have presented a multimodal model for fake news identification on social media. We combined textual features derived from a pre-trained BERT model and visual features derived from VGG-19 pre-trained on ImageNet data. Especially, we proposed a novel attention mechanism to learn the correlation between modalities. Experimental results confirmed the effectiveness of our method in the fake new detection task. In future work, we plan to use more than one image for identifying fake news. We also intend to use user-features such as number of friends, number of followers, and other metadata features from the tweets such as number of likes and comments.



\bibliography{IEEEabrv}

\begin{thebibliography}{10}
\providecommand{\url}[1]{#1}
\csname url@samestyle\endcsname
\providecommand{\newblock}{\relax}
\providecommand{\bibinfo}[2]{#2}
\providecommand{\BIBentrySTDinterwordspacing}{\spaceskip=0pt\relax}
\providecommand{\BIBentryALTinterwordstretchfactor}{4}
\providecommand{\BIBentryALTinterwordspacing}{\spaceskip=\fontdimen2\font plus
\BIBentryALTinterwordstretchfactor\fontdimen3\font minus
  \fontdimen4\font\relax}
\providecommand{\BIBforeignlanguage}[2]{{%
\expandafter\ifx\csname l@#1\endcsname\relax
\typeout{** WARNING: IEEEtranS.bst: No hyphenation pattern has been}%
\typeout{** loaded for the language `#1'. Using the pattern for}%
\typeout{** the default language instead.}%
\else
\language=\csname l@#1\endcsname
\fi
#2}}
\providecommand{\BIBdecl}{\relax}
\BIBdecl

\bibitem{Allcott16}
\BIBentryALTinterwordspacing
H.~Allcott and M.~Gentzkow, ``Social media and fake news in the 2016
  election,'' \emph{Journal of Economic Perspectives}, vol.~31, no.~2, pp.
  211--36, May 2017. [Online]. Available:
  \url{https://www.aeaweb.org/articles?id=10.1257/jep.31.2.211}
\BIBentrySTDinterwordspacing

\bibitem{ba2016layer}
J.~L. Ba, J.~R. Kiros, and G.~E. Hinton, ``Layer normalization,'' 2016.

\bibitem{baziotis-pelekis-doulkeridis:2017:SemEval2}
C.~Baziotis, N.~Pelekis, and C.~Doulkeridis, ``Datastories at semeval-2017 task
  4: Deep lstm with attention for message-level and topic-based sentiment
  analysis,'' in \emph{Proceedings of the 11th International Workshop on
  Semantic Evaluation (SemEval-2017)}.\hskip 1em plus 0.5em minus 0.4em\relax
  Vancouver, Canada: Association for Computational Linguistics, August 2017,
  pp. 747--754.

\bibitem{mediaeval16}
C.~Boididou, K.~Andreadou, S.~Papadopoulos, D.-T. Dang-Nguyen, G.~Boato,
  M.~Riegler, Y.~Kompatsiaris \emph{et~al.}, ``Verifying multimedia use at
  mediaeval 2015.'' \emph{MediaEval}, vol.~3, no.~3, p.~7, 2015.

\bibitem{Castillo2011}
\BIBentryALTinterwordspacing
C.~Castillo, M.~Mendoza, and B.~Poblete, ``Information credibility on
  twitter,'' in \emph{Proceedings of the 20th International Conference on World
  Wide Web}, ser. WWW '11.\hskip 1em plus 0.5em minus 0.4em\relax New York, NY,
  USA: Association for Computing Machinery, 2011, p. 675–684. [Online].
  Available: \url{https://doi.org/10.1145/1963405.1963500}
\BIBentrySTDinterwordspacing

\bibitem{cheema2021tibs}
G.~S. Cheema, S.~Hakimov, and R.~Ewerth, ``Tib's visual analytics group at
  mediaeval '20: Detecting fake news on corona virus and 5g conspiracy,'' 2021.

\bibitem{conneau2017deep}
A.~Conneau, H.~Schwenk, L.~Barrault, and Y.~Lecun, ``Very deep convolutional
  networks for text classification,'' 2017.

\bibitem{boididou2018detection}
Detection and visualization of misleading content~on Twitter, ``Boididou,
  christina and papadopoulos, symeon and zampoglou, markos and apostolidis,
  lazaros and papadopoulou, olga and kompatsiaris, yiannis,''
  \emph{International Journal of Multimedia Information Retrieval}, vol.~7,
  no.~1, pp. 71--86, 2018.

\bibitem{devlin2019bert}
\BIBentryALTinterwordspacing
J.~Devlin, M.-W. Chang, K.~Lee, and K.~Toutanova, ``{BERT}: Pre-training of
  deep bidirectional transformers for language understanding,'' in
  \emph{Proceedings of the 2019 Conference of the North {A}merican Chapter of
  the Association for Computational Linguistics: Human Language Technologies,
  Volume 1 (Long and Short Papers)}.\hskip 1em plus 0.5em minus 0.4em\relax
  Minneapolis, Minnesota: Association for Computational Linguistics, Jun. 2019,
  pp. 4171--4186. [Online]. Available:
  \url{https://www.aclweb.org/anthology/N19-1423}
\BIBentrySTDinterwordspacing

\bibitem{duan2020rmit}
\BIBentryALTinterwordspacing
X.~Duan, E.~Naghizade, D.~Spina, and X.~Zhang, ``{RMIT at PAN-CLEF 2020:
  Profiling Fake News Spreaders on Twitter},'' in \emph{{CLEF 2020 Labs and
  Workshops, Notebook Papers}}, L.~Cappellato, C.~Eickhoff, N.~Ferro, and
  A.~N\'ev\'eol, Eds.\hskip 1em plus 0.5em minus 0.4em\relax CEUR Workshop
  Proceedings, Sep. 2020. [Online]. Available: \url{CEUR-WS.org}
\BIBentrySTDinterwordspacing

\bibitem{ghosh2018towards}
S.~Ghosh and C.~Shah, ``Towards automatic fake news classification,''
  \emph{Proceedings of the Association for Information Science and Technology},
  vol.~55, no.~1, pp. 805--807, 2018.

\bibitem{jin-att-rnn}
\BIBentryALTinterwordspacing
Z.~Jin, J.~Cao, H.~Guo, Y.~Zhang, and J.~Luo, ``Multimodal fusion with
  recurrent neural networks for rumor detection on microblogs,'' in
  \emph{Proceedings of the 25th ACM International Conference on Multimedia},
  ser. MM '17.\hskip 1em plus 0.5em minus 0.4em\relax New York, NY, USA:
  Association for Computing Machinery, 2017, p. 795–816. [Online]. Available:
  \url{https://doi.org/10.1145/3123266.3123454}
\BIBentrySTDinterwordspacing

\bibitem{kaliyar2020fndnet}
\BIBentryALTinterwordspacing
R.~K. Kaliyar, A.~Goswami, P.~Narang, and S.~Sinha, ``Fndnet – a deep
  convolutional neural network for fake news detection,'' \emph{Cogn. Syst.
  Res.}, vol.~61, no.~C, p. 32–44, Jun. 2020. [Online]. Available:
  \url{https://doi.org/10.1016/j.cogsys.2019.12.005}
\BIBentrySTDinterwordspacing

\bibitem{Khattar2019}
\BIBentryALTinterwordspacing
D.~Khattar, J.~S. Goud, M.~Gupta, and V.~Varma, ``Mvae: Multimodal variational
  autoencoder for fake news detection,'' in \emph{The World Wide Web
  Conference}, ser. WWW '19.\hskip 1em plus 0.5em minus 0.4em\relax New York,
  NY, USA: Association for Computing Machinery, 2019, p. 2915–2921. [Online].
  Available: \url{https://doi.org/10.1145/3308558.3313552}
\BIBentrySTDinterwordspacing

\bibitem{kim2014convolutional}
\BIBentryALTinterwordspacing
Y.~Kim, ``Convolutional neural networks for sentence classification,'' in
  \emph{Proceedings of the 2014 Conference on Empirical Methods in Natural
  Language Processing ({EMNLP})}.\hskip 1em plus 0.5em minus 0.4em\relax Doha,
  Qatar: Association for Computational Linguistics, Oct. 2014, pp. 1746--1751.
  [Online]. Available: \url{https://www.aclweb.org/anthology/D14-1181}
\BIBentrySTDinterwordspacing

\bibitem{8424744}
S.~{Krishnan} and M.~{Chen}, ``Identifying tweets with fake news,'' in
  \emph{2018 IEEE International Conference on Information Reuse and Integration
  (IRI)}, 2018, pp. 460--464.

\bibitem{ma-etal-2018-rumor}
\BIBentryALTinterwordspacing
J.~Ma, W.~Gao, and K.-F. Wong, ``Rumor detection on twitter with
  tree-structured recursive neural networks,'' in \emph{Proceedings of the 56th
  Annual Meeting of the Association for Computational Linguistics (Volume 1:
  Long Papers)}.\hskip 1em plus 0.5em minus 0.4em\relax Melbourne, Australia:
  Association for Computational Linguistics, Jul. 2018, pp. 1980--1989.
  [Online]. Available: \url{https://www.aclweb.org/anthology/P18-1184}
\BIBentrySTDinterwordspacing

\bibitem{bertweet}
D.~Q. Nguyen, T.~Vu, and A.~T. Nguyen, ``{BERTweet: A pre-trained language
  model for English Tweets},'' in \emph{Proceedings of the 2020 Conference on
  Empirical Methods in Natural Language Processing: System Demonstrations},
  2020, pp. 9--14.

\bibitem{shu2017}
\BIBentryALTinterwordspacing
K.~Shu, A.~Sliva, S.~Wang, J.~Tang, and H.~Liu, ``Fake news detection on social
  media: A data mining perspective,'' \emph{SIGKDD Explor. Newsl.}, vol.~19,
  no.~1, p. 22–36, Sep. 2017. [Online]. Available:
  \url{https://doi.org/10.1145/3137597.3137600}
\BIBentrySTDinterwordspacing

\bibitem{shu2019role}
K.~Shu, X.~Zhou, S.~Wang, R.~Zafarani, and H.~Liu, ``The role of user profile
  for fake news detection,'' 2019.

\bibitem{simonyan2015deep}
K.~Simonyan and A.~Zisserman, ``Very deep convolutional networks for
  large-scale image recognition,'' 2015.

\bibitem{SpotFake}
S.~{Singhal}, R.~R. {Shah}, T.~{Chakraborty}, P.~{Kumaraguru}, and S.~{Satoh},
  ``Spotfake: A multi-modal framework for fake news detection,'' in \emph{2019
  IEEE Fifth International Conference on Multimedia Big Data (BigMM)}, 2019,
  pp. 39--47.

\bibitem{JMLR:v15:srivastava14a}
\BIBentryALTinterwordspacing
N.~Srivastava, G.~Hinton, A.~Krizhevsky, I.~Sutskever, and R.~Salakhutdinov,
  ``Dropout: A simple way to prevent neural networks from overfitting,''
  \emph{Journal of Machine Learning Research}, vol.~15, no.~56, pp. 1929--1958,
  2014. [Online]. Available:
  \url{http://jmlr.org/papers/v15/srivastava14a.html}
\BIBentrySTDinterwordspacing

\bibitem{tian2020stance}
L.~Tian, X.~Zhang, Y.~Wang, and H.~Liu, ``Early detection of rumours on twitter
  via stance transfer learning,'' in \emph{Advances in Information Retrieval},
  J.~M. Jose, E.~Yilmaz, J.~Magalh{\~a}es, P.~Castells, N.~Ferro, M.~J. Silva,
  and F.~Martins, Eds.\hskip 1em plus 0.5em minus 0.4em\relax Cham: Springer
  International Publishing, 2020, pp. 575--588.

\bibitem{8508256}
L.~{Wang}, Y.~{Wang}, G.~{de Melo}, and G.~{Weikum}, ``Five shades of untruth:
  Finer-grained classification of fake news,'' in \emph{2018 IEEE/ACM
  International Conference on Advances in Social Networks Analysis and Mining
  (ASONAM)}, 2018, pp. 593--594.

\bibitem{Wang2018EANNEA}
Y.~Wang, F.~Ma, Z.~Jin, Y.~Yuan, G.~Xun, K.~Jha, L.~Su, and J.~Gao, ``Eann:
  Event adversarial neural networks for multi-modal fake news detection,''
  \emph{Proceedings of the 24th ACM SIGKDD International Conference on
  Knowledge Discovery \& Data Mining}, 2018.

\bibitem{Wu2015FalseRD}
K.~Wu, S.~Yang, and K.~Q. Zhu, ``False rumors detection on sina weibo by
  propagation structures,'' \emph{2015 IEEE 31st International Conference on
  Data Engineering}, pp. 651--662, 2015.

\bibitem{yang2018ticnn}
Y.~Yang, L.~Zheng, J.~Zhang, Q.~Cui, Z.~Li, and P.~S. Yu, ``Ti-cnn:
  Convolutional neural networks for fake news detection,'' 2018.

\bibitem{zhou2020safe}
X.~Zhou, J.~Wu, and R.~Zafarani, ``Safe: Similarity-aware multi-modal fake news
  detection,'' 2020.

\bibitem{zhou2019networkbased}
X.~Zhou and R.~Zafarani, ``Network-based fake news detection: A pattern-driven
  approach,'' 2019.

\end{thebibliography}
\bibliographystyle{IEEEtranS}

\end{document}